\pdfoutput=1

\documentclass[11pt]{article}

\usepackage[]{EMNLP2023}

\usepackage{times}
\usepackage{latexsym}

\usepackage[T1]{fontenc}

\usepackage[utf8]{inputenc}

\usepackage{microtype}

\usepackage{inconsolata}

\usepackage{graphicx}
\usepackage{caption}
\usepackage{subcaption}
\usepackage{amsmath}
\usepackage{booktabs}
\usepackage{graphicx}
\usepackage{color,soul}
\usepackage{xcolor}
\usepackage{pifont}
\usepackage{amssymb}
\usepackage{mathtools}
\usepackage[colorinlistoftodos]{todonotes}
\usepackage{booktabs}
\usepackage{multirow}
\usepackage{float}
\usepackage{xspace}

\def\eg{{\em e.g., }\xspace}
\def\ie{{\em i.e., }\xspace}

\usepackage[capitalize,noabbrev]{cleveref}

\newcommand{\hlc}[2][yellow]{{
 \colorlet{foo}{#1}
 \sethlcolor{foo}
\hl{#2}
}}

\title{\textsc{Refer}: An End-to-end Rationale Extraction Framework \\ for Explanation Regularization}

\author{Mohammad Reza Ghasemi Madani \\
  University of Bologna\\
  \texttt{mohammadreza.ghasemi@studio.unibo.it} \\\And
  Pasquale Minervini \\
  University of Edinburgh\\
  \texttt{p.minervini@ed.ac.uk} \\}

\begin{document}
\maketitle
\begin{abstract}
%
Human-annotated textual explanations are becoming increasingly important in Explainable Natural Language Processing.
%
%
\emph{Rationale extraction} aims to provide \emph{faithful} (\ie{reflective of the behavior of the model}) and \emph{plausible} (\ie{convincing to humans}) explanations by highlighting the inputs that had the largest impact on the prediction 
without compromising the performance of the task model.
%
%
In recent works, the focus of training rationale extractors was primarily on optimizing for plausibility using human highlights, 
while the task model was trained on jointly optimizing for task predictive accuracy and faithfulness.
%
We propose \textbf{\textsc{Refer}}, a framework that employs 
a differentiable rationale extractor that allows to back-propagate through the rationale extraction process.
%
%
We analyze the impact of using human highlights during training by jointly training the task model and the rationale extractor. 
%
%
In our experiments, \textsc{Refer} yields significantly better results in terms of faithfulness, plausibility, and downstream task accuracy on both in-distribution and out-of-distribution data. 
%
%
On both e-SNLI and CoS-E, our best setting produces better results in terms of composite normalized relative gain than the previous baselines by 11\% and 3\%, respectively. 
%

\end{abstract}

\section{Introduction}
%
Neural Language Models have emerged as State-of-The-Art (SoTA) performers in a wide range of Natural Language Processing (NLP) tasks \cite{devlin-etal-2019-bert,liu2019roberta}.
However, they are often perceived as opaque \cite{rudin2019stop,doshivelez2017rigorous,lipton2018mythos}, sparking significant interest in the development of algorithms that can automatically explain the behavior of these models \cite{denil2015extraction, sundararajan2017axiomatic, Camburu_NEURIPS2018, rajani-etal-2019-explain, luo2022local}. 

In the field of self-explainable neural models, two prominent approaches have emerged: (i) Extractive Rationales~\cite[ERs,][]{zaidan-etal-2007-using, bastings-filippova-2020-elephant}, which involve selecting a subset of input features responsible for a prediction, and (ii) Natural Language Explanations \cite[NLEs,][]{DBLP:conf/cvpr/ParkHARSDR18, DBLP:conf/eccv/HendricksARDSD16, DBLP:conf/iccv/KayserCSEDAL21, Camburu_NEURIPS2018}, which generate human-readable justifications for predictions. 
The key aspects of interest for both ERs and NLEs are \emph{plausibility}, which measures the alignment between model explanations and ground truth, and \emph{faithfulness}, which measures how accurately the explanations reflect the decision-making process of the model.
ERs offer concise explanations, serving as a means for users to assess the trustworthiness of a model. However, ERs may lack important reasoning details, such as feature relationships~\cite{wiegreffe-etal-2021-measuring}.
On the other hand, NLEs provide detailed justifications in natural language, complementing ERs by 
potentially offering more comprehensive explanations. 

The evaluation of ERs involves assessing their \emph{plausibility} and \emph{faithfulness}.
Plausibility refers to the extent to which a highlight explains a predicted label, as judged by human evaluators, or according to the similarity with gold highlights \cite{yang-etal-2020-generating, deyoung-etal-2020-eraser}. 
Faithfulness measures how accurately a highlight represents the decision process of the model -- for example, by measuring to which extent the confidence in the predicted label changes after removing the highlighted words (\emph{comprehensiveness}) or when only considering the highlighted words (\emph{sufficiency}) \cite{Melis_NEURIPS2018_3e9f0fc9, wiegreffe-pinter-2019-attention}.
%
%
%

%
Previous works largely focused on rationale extraction, which involves explaining the output of a model by identifying the input tokens that exert the greatest influence on model predictions \cite{denil2015extraction,sundararajan2017axiomatic,jin2020hierarchical,lundberg2017unified} and 
providing additional supervision signal \cite{hase-bansal-2022-models}.
%
%
%
%
The majority of prior works in this area have revolved around \emph{explanation regularization}, a technique aimed at improving generalization in neural models by aligning machine rationales with human rationales \cite{ross2017right,huang-etal-2021-exploring,ghaeini-etal-2019-saliency,kennedy-etal-2020-contextualizing,rieger2020interpretations,liu-avci-2019-incorporating}.
However, ERs are discrete distributions over the input text, which can be difficult to learn by neural models via back-propagation~\cite{niepert2021implicit}.
%
%
In this work, we propose \textsc{\textbf{Refer}}, an End-to-end \textbf{R}ationale \textbf{E}xtraction \textbf{F}ramework for \textbf{E}xplanation \textbf{R}egularization, which allows to back-propagate through the rationale extraction process. 
Specifically, \textsc{Refer} involves a differentiable rationale extractor, which selects the top-$k\%$ 
most important words from the textual input, which are then used by the model to generate a prediction. 

\begin{figure}[t]
    \centering
    \includegraphics[width=\columnwidth]{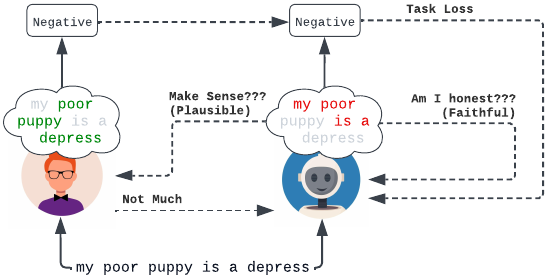}
    \caption{Explanation Regularization System: model is trained with human rationales while maintaining high task performance. In this case, the model predicts the correct label for incorrect reasons.} 
    \label{fig:expl-reg}
\end{figure}

\section{Related Works}
%
%
%
The inherent complexity of neural models has given rise to concerns regarding their opacity \cite{rudin2019stop}, particularly about the societal implications of employing neural models in high-stakes decision-making scenarios \cite{bender2021dangers}.
Therefore, 
explainability is of utmost importance for fostering trust, ensuring ethical practices, and maintaining the safety of NLP systems \cite{doshivelez2017rigorous,lipton2018mythos}. 

%

\paragraph{Learning to Explain}
%
Rationalization 
offers local explanations by providing a unique explanation for each prediction instead of a global explanation that covers the entire model \cite{JMLR:v11:baehrens10a, ribeiro-etal-2016-trust}.
These explanations yield valuable insights for various purposes, including debugging, quantifying bias and fairness, understanding model behavior, and ensuring robustness and privacy \cite{molnar2022}.
However, obtaining direct supervision in the form of human-labeled rationales during training is not always feasible, which has led to the development of datasets that include human justifications for the true labels. 
%
%
These efforts enhance the interpretability of NLP models and address the limitations associated with direct supervision in learning to explain.
\paragraph{Post-hoc Explanations}
%
%
Post-hoc explanations are another branch of interpretability research.
These explanations often involve token-level importance scores. In the quest for effective post-hoc explanations, a balance must be struck between the clarity of semantics and the avoidance of counter-intuitive behaviors.
Gradient-based explanations \cite{sundararajan2017axiomatic, smilkov2017smoothgrad} provide clear semantics by describing the local impact of input perturbations on the outputs of the model.
However, they can sometimes exhibit inconsistent behaviors \cite{feng-etal-2018-pathologies}, and their effectiveness relies on the differentiability of the model.
Alternatively, there are model-agnostic methods that do not rely on specific model properties.
One notable example is Local Interpretable Model-agnostic Explanations \cite[LIME,][]{ribeiro-etal-2016-trust}.
These approaches approximate the behavior of the model locally by repeatedly making predictions on perturbed inputs and fitting a simple, explainable model over the resulting outputs.
%
%

%
\paragraph{Learning from Human Rationales}
%
%
Recent research has focused on leveraging rationales to enhance the training of neural text classifiers.
\citet{zhang-etal-2016-rationale} introduced a rationale-augmented Convolutional Neural Network that explicitly identifies sentences supporting categorizations.
\citet{strout-etal-2019-human} demonstrated that incorporating rationales during training improves the quality of predicted rationales, as preferred by humans compared to models trained without explicit supervision \cite{strout-etal-2019-human}.
In addition to integrated models, pipeline approaches have been proposed, where separate models are trained for rationale extraction and classification based on these extracted rationales \cite{lehman-etal-2019-inferring, chen-etal-2019-seeing}.
These approaches assume the availability of explicit training data for rationale extraction.
%

\begin{figure}[t]
    \centering
    \includegraphics[width=\columnwidth]{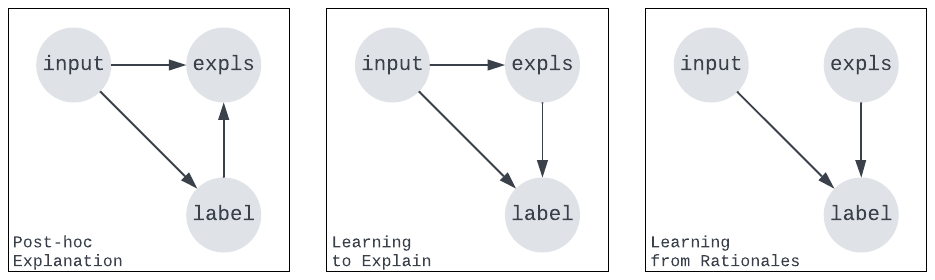}
    \caption{
Computation graphs describing the relationships between post-hoc explanations, learning to explain, and learning from rationales.}
    \label{fig:expl-types}
\end{figure}
%
%
\paragraph{Extractive Rationale Objectives} Several prior works have aimed to enhance the \emph{faithfulness} of extractive rationales using Attribution Algorithms (AAs), which extract rationales via handcrafted functions~\cite{sundararajan2017axiomatic,ismail2021nips,situ-etal-2021-learning}. 
%
%
However, AAs are not easily optimized and often require significant computational resources. 
\citet{situ-etal-2021-learning,schwarzenberg-etal-2021-efficient} tackle the computational cost by training a model 
to mimic the behavior of an AA.
\citet{jain-etal-2020-learning, Yu_NEURIPS2021_6a711a11, paranjape-etal-2020-information, bastings-filippova-2020-elephant, yu-etal-2019-rethinking, lei-etal-2016-rationalizing} use Select-Predict Pipelines (SPPs) to generate faithful rationales. However, SPPs only guarantee sufficiency but not comprehensiveness~\cite{deyoung-etal-2020-eraser}, and generally produce less accurate results, 
since they can only observe a portion of the input, 
and due to the challenges associated with gradient-based optimization and discrete distributions.

Regarding the \emph{plausibility} of the rationales, existing approaches typically involve supervising neural rationale extractors~\cite{bhat-etal-2021-self} and SPPs~\cite{jain-etal-2020-learning, paranjape-etal-2020-information, deyoung-etal-2020-eraser} using gold rationales. However, LM-based extractors lack training for faithfulness, and SPPs sacrifice task performance to achieve faithfulness by construction. Other works mainly focus on improving the plausibility of rationales \cite{narang2020wt5, lakhotia-etal-2021-fid, Camburu_NEURIPS2018}, often employing task-specific pipelines \cite{rajani-etal-2019-explain, kumar-talukdar-2020-nile}.
In contrast, \textsc{Refer} \emph{jointly} optimizes both the task model and rationale extractor for faithfulness, plausibility, and task performance and reaches a better trade-off w.r.t. these desiderata without suffering from heuristic-based approaches (\eg{AAs}) disadvantages.
%
%
%

\section{Model Architecture}
\paragraph{Task Model}
Consider \(\mathcal{F}_{\text{task}}\) as the task model for text classification, where it consists of an encoder \cite{vaswani2017att} and a head. 
%
Let \(\text{x}_i = [\text{x}^t_i]^n_{t=1}\) be $i^{th}$ input sequence with length $n$, and \(\mathcal{F}_{\text{task}}(\text{x}_i) \in \mathbb{R}^M\) be the logit vector for the output of the task model.
We use \(y_i = \arg\max_j [\mathcal{F}_{\text{task}}(\text{x}_i)]_{j} \) to denote the class predicted by task model. Given that cross-entropy loss is used to train \(\mathcal{F}_{\text{task}}\) to predict $y_i^{*}$, the task loss is defined as follow:
\begin{equation}
    \mathcal{L}_{\text{task}} = \mathcal{L}_{\text{CE}} (\mathcal{F}_{\text{task}}(\text{x}_i), y_i^{*})
\end{equation}
\begin{figure}[t]
    \centering
    \includegraphics[width=\columnwidth]{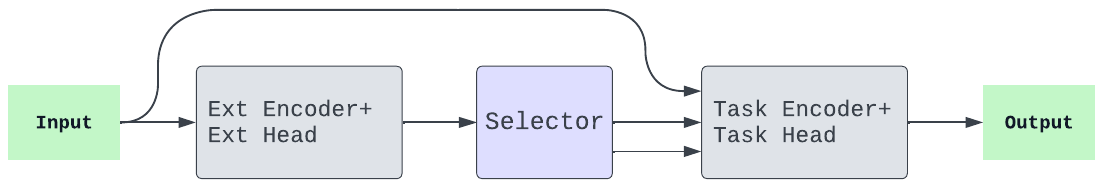}
    \caption{The pipeline for explanation regularization is a fully end-to-end approach where the task model's output loss is back-propagated through all components, resulting in a compromised performance that considers all training criteria.}
    \label{fig:our-pipeline}
\end{figure}
\paragraph{Rationale Extractor}
Let \(\mathcal{F}_{\text{ext}}\) denote a rationale extractor, such that \(\text{s}_i = \mathcal{F}_{\text{ext}}(\text{x}_i)\). 
Given \(\mathcal{F}_{\text{task}}\), \(\text{x}_i\), and $y_i$, the goal of rationale extraction is to output vector \(\text{s}_i = [s^t_i]^n_{t=1} \in \mathbb{R}^{n}\), such that each \(s^t_i\) is an importance score indicating how strongly token $\text{x}^t_i$ influenced \(\mathcal{F}_{\text{task}}\) to predict class $y_i$.
The final rationales are typically obtained by binarizing $\text{s}_i$ as \(\text{r}_i^{(k)} \in \{0, 1\}^n\), via the top-$k\%$ strategy 
\cite{deyoung-etal-2020-eraser, jain-etal-2020-learning, pruthi-etal-2022-evaluating, chan_NEURIPS2021_9752d873}. 
%
%
%
%

%
To capture the degree to which the snippets within the extracted rationales are sufficient for a model to make a prediction, we measure the disparity in model confidence when considering the complete input versus only the extracted rationales. 
A small difference suggests 
the high importance of extracted rationales.
\begin{equation}
\label{eqn:suff-diff}
\begin{aligned}
\mathcal{L}_{\text{suff-diff}} = \mathcal{L}_{\text{CE}} (\mathcal{F}_{\text{task}}(\text{r}_i^{(k)}), y_i^{*}) \\
- \mathcal{L}_{\text{CE}} (\mathcal{F}_{\text{task}}(\text{x}_i), y_i^{*})
\end{aligned}
\end{equation}
Following \citet{chan-etal-2022-unirex}, to avoid negative losses, we can use margin $m_s$ to impose a lower bound on $\mathcal{L}_{\text{suff-diff}}$, yielding the following margin criterion:
\begin{equation}
    \mathcal{L}_{\text{suff}} = \max(-m_s,\mathcal{L}_{\text{suff-diff}}) + m_s
\end{equation}
To compute comprehensiveness we create contrast examples for $\text{x}_i$, $\tilde{\text{x}}_i=\text{x}_i \backslash \text{r}_i^{(k)}$, which is $\text{x}_i$ with the predicted rationales $\text{r}_i$ removed \cite{zaidan-etal-2007-using}. Similar to \cref{eqn:suff-diff}, we measure the difference in model confidence between considering the complete input and the contrast set $\tilde{\text{x}}_i$.
A high score here implies that the rationales were influential in the prediction.
\begin{equation}
\begin{aligned}
\mathcal{L}_{\text{comp-diff}} = \mathcal{L}_{\text{CE}} (\mathcal{F}_{\text{task}}(\text{x}_i), y_i^{*}) \\
- \mathcal{L}_{\text{CE}} (\mathcal{F}_{\text{task}}(\tilde{\text{x}}_i), y_i^{*})
\end{aligned}
\end{equation}
%
Repeatedly, we enforce $\mathcal{L}_{\text{comp-diff}}$ to be positive as follows: 
\begin{equation}
    \mathcal{L}_{\text{comp}} = \max(-m_c,\mathcal{L}_{\text{comp-diff}}) + m_c
\end{equation}
%
Finally, the selection of the tokens 
for matching the human highlights can be cast as a binary classification problem,
%
and the plausibility loss is computed using the binary cross-entropy (BCE) loss function:
\begin{equation}
    \mathcal{L}_{\text{plaus}} = - \sum_t \text{r}_i^{*,t} \log(\mathcal{F}_{\text{ext}}(\text{x}_i^{t}))
\end{equation}
where $\text{r}_i^{*}$ is the gold rationale for input $\text{x}_i$ of length $t$. This leads to the following 
multi-task learning objective:
\begin{equation*}
\begin{aligned}
    \mathcal{L} & = \mathcal{L}_{\text{task}} + \alpha_{\text{f}}\mathcal{L}_{\text{faith}} + \alpha_{\text{p}}\mathcal{L}_{\text{plaus}} \\
    & = \mathcal{L}_{\text{task}} + \alpha_{\text{c}}\mathcal{L}_{\text{comp, K}} + \alpha_{\text{s}}\mathcal{L}_{\text{suff, K}} + \alpha_{\text{p}}\mathcal{L}_{\text{plaus}}
\end{aligned}
\end{equation*}
\paragraph{Back-Propagating Through Rationale Extraction}
%
%
%
%
%
%
%
%
%
%
%
To back-propagate through the rationale extraction process, we use Adaptive Implicit Maximum Likelihood Estimation~\citep[AIMLE,][]{aimle23}, a recently proposed low-variance and low-bias gradient estimation method for discrete distribution that does not require significant hyper-parameter tuning.
AIMLE is an extension of Implicit Maximum Likelihood Estimation~\citep[IMLE,][]{niepert2021implicit}, a perturbation-based gradient estimator where the gradient of the loss w.r.t. the token scores $\nabla_{\mathbf{s}} \mathcal{L}$ is estimated as $\nabla_{\mathbf{s}} \mathcal{L} \approx \mathbf{r}(\mathbf{s} + \epsilon) - \mathbf{r}(\mathbf{s} + \lambda \nabla_{\mathbf{r}} \mathcal{L} + \epsilon)$, where $\epsilon$ denotes Gumbel noise, $\mathbf{r}$ denotes the top-$k$\% function, and $\lambda$ is a hyper-parameter selected by the user.
AIMLE removes the need for the user to select $\lambda$ by automatically identifying the optimal $\lambda$ for a given learning task.
%
%
%
%
%

\begin{figure}[t]
    \centering
    \includegraphics[width=\columnwidth]{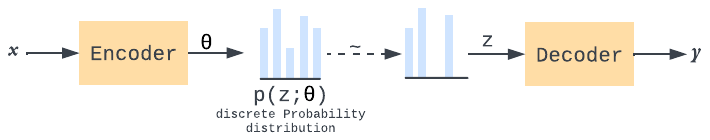}
    \caption{Illustration of the learning problem. $z$ is the discrete latent structure, $x$ and $y$ are feature inputs and target outputs, Encoder maps $\mathcal{X} \mapsto \theta$, Decoder maps $\mathcal{Z} \mapsto \mathcal{Y}$, and $p(z; \theta)$ represents the discrete probability distribution. The dashed path indicates non-differentiability.} 
    \label{fig:discrete-latent-space}
\end{figure}


\begin{figure*}[t]
    \centering
    \includegraphics[width=\textwidth]{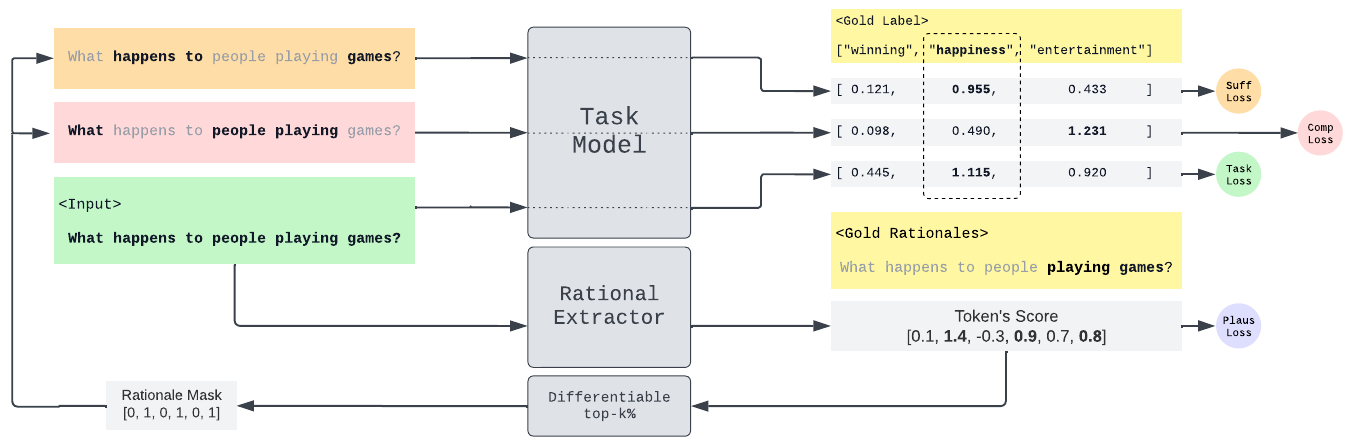}
    \caption{\textbf{\textsc{Refer} Pipeline}. The Task Model is trained using (i) Task Loss, (ii) Sufficiency Loss, and (iii) Comprehensiveness Loss, while the Rationale Extractor is trained through backpropagation using (i) Plausibility Loss, (ii) Sufficiency Loss, and (iii) Comprehensiveness Loss. This approach ensures a high level of consistency across each criterion, as all components are aware of each other's status and can adapt to strike a balance among the three criteria.} 
    \label{fig:detailed-pipeline}
\end{figure*}

\section{Research Questions} 
%

%
\paragraph{\textit{RQ1: Does training the model on human highlights improve the generalization properties of the model?}}
Nowadays, machine learning systems 
can learn to capture spurious correlations in the data for solving any given task, and often struggle in more challenging cases \cite{mccoy-etal-2019-right}. 
When models are allowed to make predictions without considering rationales-related criteria---faithfulness and plausibility---the rationales extracted by the model can be incomprehensible and lack meaningful interpretations~\cite{vig-belinkov-2019-analyzing}. 
%
%
%
%
Without understanding the factors and information that influence the predictions of the model, it becomes difficult to trust or explain its outputs.
%
%
%
In certain contexts, faithful explanations are crucial -- for example, 
they can be used to determine whether a model relies on protected attributes, such as gender or religious group~\cite{pruthi-etal-2020-learning}.
\citet{mccoy-etal-2019-right} propose the hypothesis that neural natural language inference (NLI) models might rely on three fallible syntactic heuristics: (i) lexical overlap, (ii) subsequences, and (iii) constituents. 
%
%
To evaluate whether the models have indeed adopted these heuristics, we use Heuristic Analysis for NLI Systems \citep[HANS,][]{mccoy-etal-2019-right}, which includes a variety of examples where such heuristics fail, providing a means to assess a model's reliance on these heuristics. \cref{tab:huristics} shows instances of these heuristics in the HANS dataset.
%
%
%

%
Faithfulness refers to the degree to which an explanation provided by a model accurately reflects the information utilized by the model to make a decision~\citep{jacovi-goldberg-2020-towards}. 
%
%
%
%
they can be used to determine whether a model is relying on protected attributes, such as gender or religious group~\cite{pruthi-etal-2020-learning}.
%
%
%
%
\paragraph{\textit{RQ2: How can we make machines imitate human rationales?}}
Human rationales are often derived from their extensive background knowledge and understanding of various concepts. While language models (LMs) possess some degree of this knowledge, they face challenges in balancing between optimizing for task performance and meeting the criteria for extractive explanations. Therefore, balancing plausibility, faithfulness, and task accuracy presents a challenging task.
%
%
%
%
%
A model can reflect its inner process to make a prediction (faithful), but it may not make sense for humans (implausible).
On the other hand, a model that returns convincing rationales (plausible) without using them during decision-making is not very useful (unfaithful).
%
%

\paragraph{\textit{RQ3: Does training the model on a small number of human highlights improve its generalization properties?}} 
Humans can efficiently learn new tasks with only a few examples by leveraging their prior knowledge. 
Recent approaches for rationalizing rely on a large number of labeled training examples, including task labels and annotated rationales for each instance. 
Obtaining such extensive annotations is often infeasible for many tasks. 
Additionally, fine-tuning LMs, which typically have billions of parameters, can be expensive and prone to overfitting. 
%
%
%
Given the high cost of human annotations, a more practical approach involves incorporating a limited amount of human supervision. 
We investigate the characteristics of effective rationales and demonstrate that making the neural model aware of its rationalized predictions can significantly enhance its performance, especially in low-resource scenarios.
%
%
%
%
\begin{figure*}[t]
    \centering
    \includegraphics[width=\textwidth]{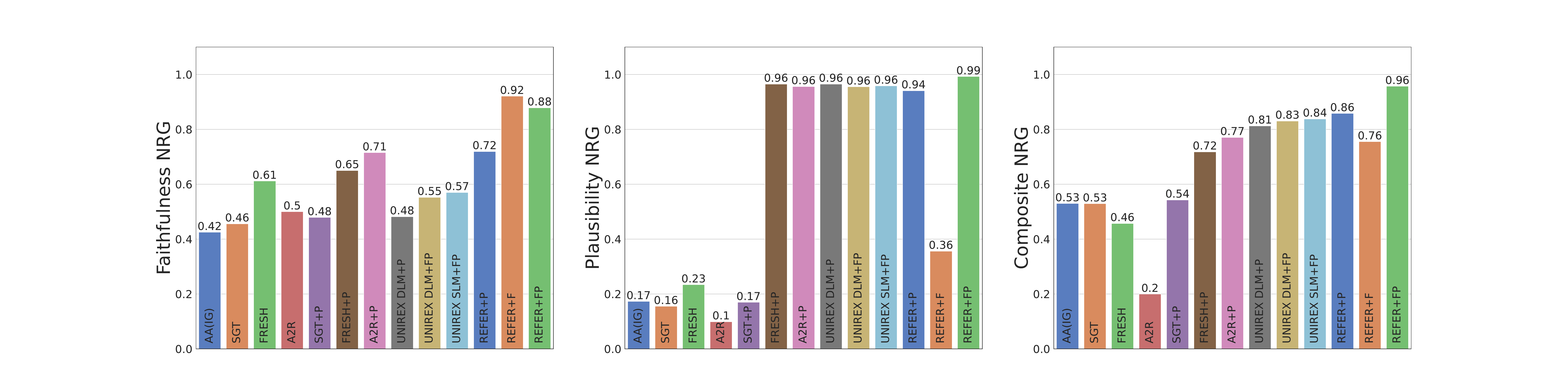}
    \caption{Comparison of models w.r.t. faithfulness NRG (FNRG), plausibility NRG (PNRG), and composite NRG (CNRG). +P, +F, +FP indicate whether the model was regularized for plausibility, faithfulness, or both.} 
    \label{fig:esnli-results}
\end{figure*}
\paragraph{\textit{RQ4: Do the learned rationale extractors generalize to OOD data?}}
%
%
The poor performance of models on OOD datasets can stem from limitations in the model's architecture, insufficient signals in the OOD training set, or a combination of both \cite{mccoy-etal-2019-right}. 
An NLI system that correctly labels an example may not do so by understanding the meaning of the sentences but rather by relying on the assumption that any hypothesis with words appearing in the premise is entailed by the premise \cite{DBLP:conf/cogsci/DasguptaGSGG18, naik-etal-2018-stress}.
\citet{gururangan-etal-2018-annotation} raises doubts about whether models trained on the SNLI dataset truly learn language comprehension or primarily rely on spurious correlations, also known as artifacts. 
%
%
For instance, words like "friends" and "old" frequently appear in neutral hypotheses. 
%
%
%
%
%
%
To analyze this, we evaluate our model on contrast sets \cite{gardner-etal-2020-evaluating} as well as unseen data, which are (mostly) label-changing small perturbations on instances to understand the true local boundary of the dataset. Essentially, they help us understand if the rationale extractor has learned any dataset-specific shortcuts. \cref{tab:mnli-contrast} shows samples for both label-changing and and non-label-changing instances modified by \citet{li-etal-2020-linguistically}.
\section{Experiment}

\subsection{Baselines}
The first class of baselines is AAs, which do not involve training \(\mathcal{F}_{\text{ext}}\) and is applied post hoc (i.e., they do not impact \(\mathcal{F}_{\text{task}}\)’s training). Integrated Gradient baseline~\cite[AA (IG),][]{sundararajan2017axiomatic} is utilized as a baseline for this class. Saliency Guided Training \cite[SGT,][]{ismail2021nips} is another baseline that uses a sufficiency-based criterion to regularize \(\mathcal{F}_{\text{task}}\), such that the AA yields faithful rationales for \(\mathcal{F}_{\text{task}}\).

Another approach is the Select-Predict Pipeline (SPP), wherein \(\mathcal{F}_{\text{task}}\) is trained to solve a given task using only the tokens chosen by \(\mathcal{F}_{\text{ext}}\) \cite{jain-etal-2020-learning, yu-etal-2019-rethinking, paranjape-etal-2020-information}; therefore, SPPs aim for "faithfulness by construction". FRESH~\cite{jain-etal-2020-learning} and A2R~\cite{yu-etal-2019-rethinking} have been proposed to produce faithful rationales: FRESH relies on training \(\mathcal{F}_{\text{task}}\) and \(\mathcal{F}_{\text{ext}}\) separately, while A2R aims to improve \(\mathcal{F}_{\text{task}}\)'s task performance by regularizing it with an attention-based predictor that utilizes the full input \cite{jain-etal-2020-learning, yu-etal-2019-rethinking}.

The most recent pipeline is UNIREX \cite{chan-etal-2022-unirex}, which considers two main architecture variants: 
(i) Dual LM (DLM), where \(\mathcal{F}_{\text{task}}\) and \(\mathcal{F}_{\text{ext}}\) are two separate Transformer-based LMs with the same encoder architecture (ii) Shared LM (SLM), where \(\mathcal{F}_{\text{task}}\) and \(\mathcal{F}_{\text{ext}}\) share encoder, while \(\mathcal{F}_{\text{ext}}\) has its own output head. \cref{fig:slm-dlm} shows the architecture for DLM and SLM in UNIREX. 
DLM provides more capacity for \(\mathcal{F}_{\text{ext}}\), which can help \(\mathcal{F}_{\text{ext}}\) provide plausible rationales. While SLM leverages multitask learning and improve faithfulness since \(\mathcal{F}_{\text{ext}}\) has greater access to information about \(\mathcal{F}_{\text{task}}\)’s reasoning process \cite{chan-etal-2022-unirex}. 
\textsc{Refer} benefits from both SLM and DLM architectures by establishing communication between separate \(\mathcal{F}_{\text{task}}\) and \(\mathcal{F}_{\text{ext}}\) using back-propagation. 

\subsection{Metrics}
To evaluate faithfulness, plausibility, and task performance, we adopt the metrics established in the ERASER benchmark \cite{deyoung-etal-2020-eraser} and UNIREX \cite{chan-etal-2022-unirex}. For assessing faithfulness, we use \emph{comprehensiveness} and \emph{sufficiency}, 
and calculate the final comprehensiveness and sufficiency metrics using the area-over-precision curve (AOPC).
%
Measuring exact matches between predicted and reference rationales is likely too strict; thus, \citet{deyoung-etal-2020-eraser} also consider the Intersection-Over-Union (IOU) 
which permits credit assignment for partial matches. 
We use these partial matches to calculate the Area Under the Precision-Recall Curve (AUPRC) and Token F1 (TF1) to quantify the similarity between the extracted rationales and the gold rationales \cite{deyoung-etal-2020-eraser, narang2020wt5}.
Also, we use accuracy and macro F1 to evaluate the task model performance on CoS-E and e-SNLI, respectively.
%
To compare different methods w.r.t. all three desiderata, \citet{chan-etal-2022-unirex} utilized the Normalized Relative Gain (NRG) metric 
that maps all raw scores to range $[0, 1]$ --- the higher the better. 
%
%
%
%
%
Finally, to summarize all of the raw metrics, we compute 
single NRG score by averaging the NRG scores for faithfulness, plausibility, and task accuracy. 
%
%

\subsection{Datasets}


We primarily experiment with the CoS-E \cite{rajani-etal-2019-explain} and e-SNLI \cite{Camburu_NEURIPS2018} datasets, all of which have gold rationale annotations from ERASER \cite{deyoung-etal-2020-eraser}. For the OOD generalization evaluation, we consider MNLI \cite{williams-etal-2018-broad} and HANS \cite{mccoy-etal-2019-right}.

\paragraph{CoS-E}~\cite{rajani-etal-2019-explain} consists of multiple-choice questions and answers taken from the work of \cite{talmor-etal-2019-commonsenseqa}. It includes supporting rationales for each question-answer pair in two forms. 
Extracted supporting snippets 
and free-text descriptions that provide a more detailed explanation of the reasoning behind the answer choice. 


\paragraph{e-SNLI}~\cite{Camburu_NEURIPS2018} is an augmentation of the SNLI corpus \cite{bowman-etal-2015-large} and includes human rationales as well as natural language explanations. 
For neutral pairs, annotators could only highlight words in the hypothesis. 
Furthermore, they consider explanations involving contradiction or neutrality to be correct as long as at least one piece of evidence in the input is highlighted.
Focusing on the hypothesis and allowing partial highlighting of evidence leads to the collection of non-comprehensive highlights in the dataset.
\paragraph{MNLI}~\cite{williams-etal-2018-broad} 
covers a broader range of written and spoken text, subjects, styles, and levels of formality compared to SNLI. It was introduced to determine the logical relationship between two given sentences. 
To evaluate the plausibility metrics on OOD data, we performed a random sampling of 50 instances from the MNLI validation split and annotated them manually w.r.t. gold labels. We referred to this particular subset of data as \textbf{e-MNLI}. 
\cref{tab:emnli} shows instances from e-MNLI for different labels. 
To conduct additional OOD generalization evaluation, we utilized two OOD Contrast Sets called \textbf{MNLI-Contrast} and \textbf{MNLI-Original}. These contrast sets were created by slightly modifying the original MNLI instances \cite{li-etal-2020-linguistically}. In MNLI-Contrast, the modification changes the original label, while in MNLI-Original, the original label remains the same. Examples of these contrast sets are shown in \cref{tab:mnli-contrast}.
\paragraph{HANS}~\cite{mccoy-etal-2019-right} is designed to evaluate the capability of NLI systems to rely on heuristics and patterns instead of genuine understanding. 
HANS consists of sentence pairs carefully crafted to mislead models using three heuristic categories: Lexical Overlap, Subsequence, and Constituent. Instances for each heuristic are given in \cref{tab:huristics}. By evaluating models on the HANS dataset, researchers can gain insights into the limitations and robustness of NLI systems. 

\section{Results}
\paragraph{\textit{RQ1: Does training the model on human highlights improve the generalization properties of the model? 
}} 
We label with \texttt{+P} and \texttt{+FP} the models trained by optimizing for plausibility and jointly faithfulness and plausibility, respectively.
\cref{fig:esnli-results} displays the main results for e-SNLI in terms of NRG.
Overall, \texttt{REFER+FP} achieved the highest composite NRG, improving over the strongest baseline (\texttt{UNIREX SLM+FP}) by 12\%. 
Regarding plausibility, models explicitly trained for plausibility (\texttt{+P}) or both faithfulness and plausibility (\texttt{+FP}) achieved similar results, with \texttt{REFER+FP} outperforming the second-best model by 3\%. 
Regarding faithfulness, \textsc{Refer} achieved the highest score in all three configurations.
An interesting finding is that even when training \textsc{Refer} and A2R solely for plausibility (\texttt{REFER+P} and \texttt{A2R+P}), their faithfulness NRG scores remain considerably higher than all other methods. 
Detailed results are shown in \cref{tab:cose-results} and \cref{tab:esnli-results}.
Additionally, we analyzed the model's predictions on correctly labeled instances compared to falsely labeled ones, as presented in Table \ref{tab:preds-devided}. Surprisingly, although the model achieves relatively high plausibility scores, the sufficiency and comprehensiveness metrics are low when the model predicts the wrong label. This suggests that even when human rationales are extracted from the inputs, the model does not strongly rely on them in falsely labeled input.
%

%
%
%
%
The extracted rationales by the model, shown in \cref{tab:highlights}, demonstrate the impact of regularization on explanation regularization. 
Without ER regularization, the model's reasoning tends to rely on specific data patterns and heuristics rather than meaningful explanations. 
%
%
%
In contrast, when the model is regularized on ER, the quality of the rationales improves significantly in terms of faithfulness and plausibility. For instance, the example highlights the selection of "man pushing cart" and "woman smoking cigarette" as rationales to predict the label contradiction. 
The evaluation metrics for faithfulness on e-SNLI in \cref{tab:ood} further support the notion that the model genuinely relies on these rationales for its predictions.
\begin{table}[t]
\centering
\caption{Comparison of ER metrics for truly predicted labels and falsely predicted labels. (↑) indicates the higher value is better and (↓) the lower is better.}
\label{tab:preds-devided}
\resizebox{\columnwidth}{!}{%
\begin{tabular}{@{}lcc@{}}
\toprule
\multicolumn{1}{c}{\textbf{Metrics}} & \textbf{True Predictions} & \textbf{Wrong Predictions} \\ \midrule
Sufficiency AOPC (↓)       & 0.0488 & 0.1566 \\
Comprehensiveness AOPC (↑) & 0.3311 & 0.3057 \\
Plausibility TF1 (↑)       & 0.8016 & 0.7012 \\
Plausibility AUPRC (↑)     & 0.8834 & 0.7350 \\ \bottomrule
\end{tabular}%
}
\end{table}
\begin{table}[t]
\centering
\caption{\textsc{Refer} highlights on e-SNLI. Instead of visualizing hard tokens selected by the model, we highlighted all the words w.r.t. their score.}
\label{tab:highlights}
\resizebox{\columnwidth}{!}{%
\begin{tabular}{@{}cl@{}}
\toprule
\textbf{Model} & \multicolumn{1}{c}{\textbf{Highlights}}              \\ \midrule
Original Instance           & 
\begin{tabular}[c]{@{}l@{}}\textbf{\textit{Premise}}: A\hlc[yellow!30]{man}in green pants and blue shirt pushing a cart.\\ 
\textbf{\textit{Hypothesis}}: A\hlc[yellow!30]{woman}is smoking a cigarette.\\ 
\textbf{\textit{Label}}: contradiction
\end{tabular} \\
               &
               \\
\begin{tabular}[c]{@{}c@{}}\textsc{Refer} without\\ER regularization\end{tabular}    & 
\begin{tabular}[c]{@{}l@{}}
\textbf{\textit{Premise}}:\hlc[orange!67]{A}\hlc[orange!72]{man}\hlc[orange!0]{in}\hlc[orange!57]{green}\hlc[orange!6]{pants}\hlc[orange!20]{and}\hlc[orange!61]{blue}\hlc[orange!40]{shirt}\hlc[orange!21]{pushing}\hlc[orange!17]{a}\hlc[orange!39]{cart}\hlc[orange!73]{.}\\ 
\textbf{\textit{Hypothesis}}:\hlc[orange!67]{A}\hlc[orange!71]{woman}\hlc[orange!31]{is}\hlc[orange!26]{smoking}\hlc[orange!22]{a}\hlc[orange!47]{cigarette}\hlc[orange!100]{.}\\ 
\textbf{\textit{Predict}}: contradiction
\end{tabular} \\
               &
               \\
\begin{tabular}[c]{@{}c@{}}\textsc{Refer} with\\ER regularization\end{tabular}    & 
\begin{tabular}[c]{@{}l@{}}
\textbf{\textit{Premise}}:\hlc[orange!40]{A}\hlc[orange!100]{man}\hlc[orange!8]{in}\hlc[orange!14]{green}\hlc[orange!14]{pants}\hlc[orange!0]{and}\hlc[orange!7]{blue}\hlc[orange!10]{shirt}\hlc[orange!70]{pushing}\hlc[orange!56]{a}\hlc[orange!61]{cart}\hlc[orange!60]{.}\\ 
\textbf{\textit{Hypothesis}}:\hlc[orange!40]{A}\hlc[orange!98]{woman}\hlc[orange!21]{is}\hlc[orange!70]{smoking}\hlc[orange!56]{a}\hlc[orange!62]{cigarette}\hlc[orange!61]{.}\\ 
\textbf{\textit{Predict}}: contradiction
\end{tabular} \\ \bottomrule
\end{tabular}%
}
\end{table}
\begin{figure}[t]
    \centering
    \includegraphics[width=\columnwidth]{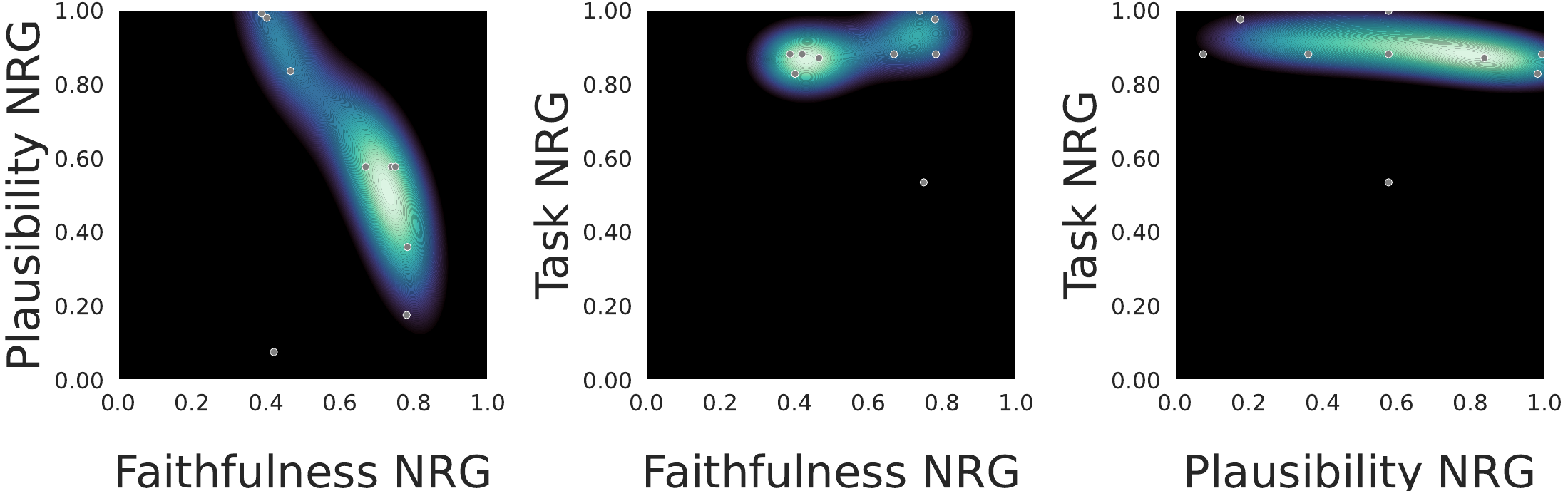}
    \caption{Results distribution of CoS-E dev split for different faithfulness and plausibility weights and $k$=50\%. Kernel Density Estimation is used to have smoothed distribution over discrete data points for visualization purposes.} 
    \label{fig:cose-dev-dist}
\end{figure}
\begin{figure}[t!]
    \centering
    \includegraphics[width=\columnwidth]{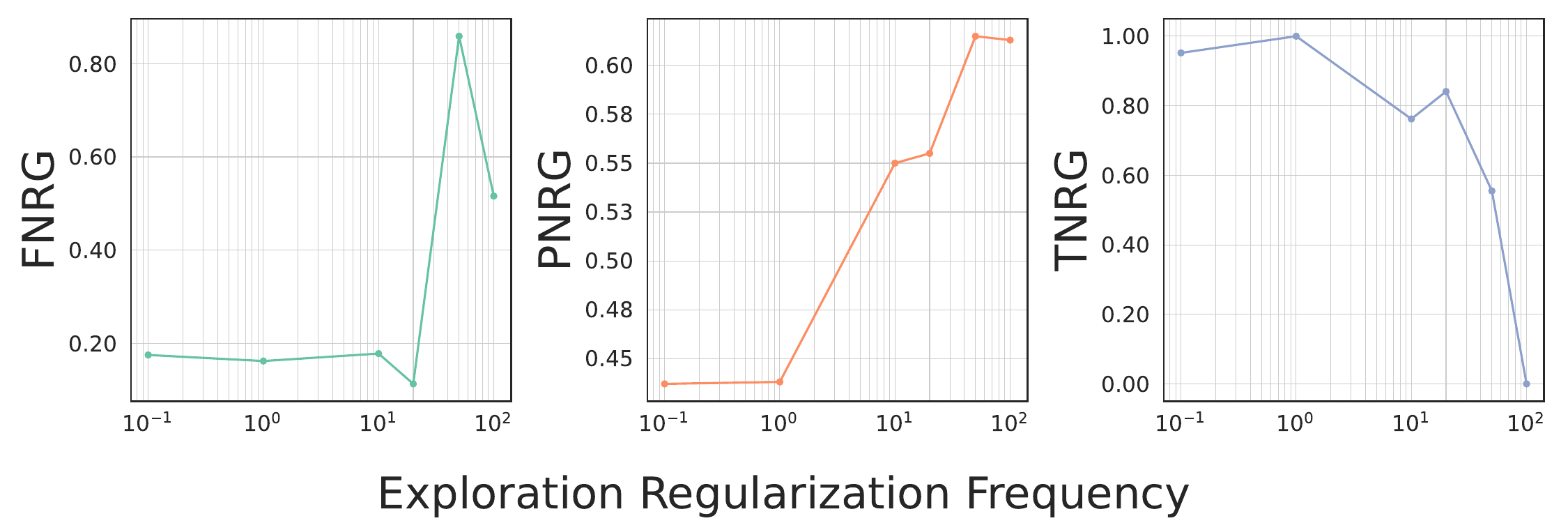}
    \caption{Comaprioson of different models w.r.t. faithfulness NRG (FNRG), plausibility NRG (PNRG), and composite NRG (CNRG).} 
    \label{fig:cose-portion}
\end{figure}
\paragraph{\textit{RQ2: How can we make machines imitate humans' rationales?}} \cref{fig:cose-dev-dist} shows the distribution of the results for different combinations of faithfulness and plausibility loss weights on the CoS-E validation set. 
We trained the model for $(\alpha_f,\alpha_p) \in \{0.0, 0.5, 1.0\}^2$. 
Based on the results, there is a slight reverse correlation between plausibility and faithfulness. 
However, the task shows relatively stable behavior over faithfulness and plausibility variation.
This means that, with our pipeline, we cannot reach a higher plausibility and faithfulness trade-off from a certain level on CoS-E.
\begin{figure}[t]
    \centering
    \includegraphics[width=\columnwidth]{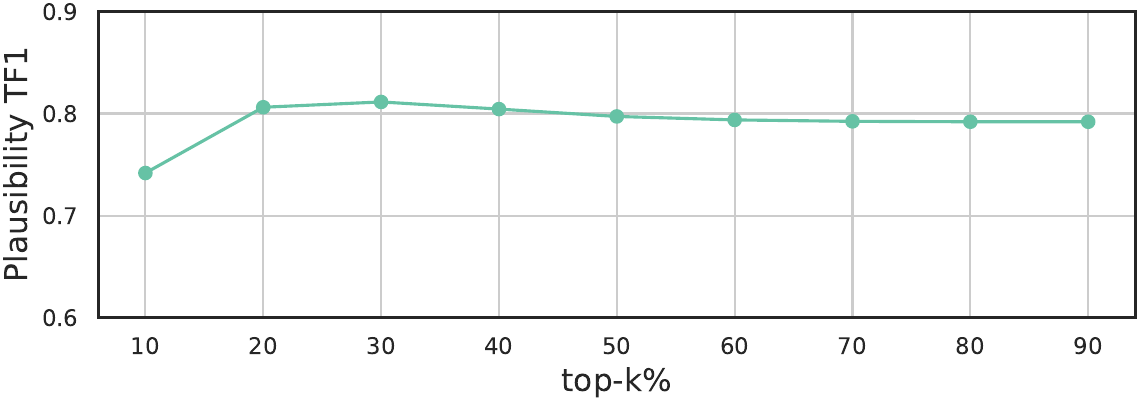}
    \caption{Plausiblity TF1 score of model trained for top-$50$\% and evaluated for other top-$k$\%s.} 
    \label{fig:topk-plaus}
\end{figure}
\paragraph{\textit{RQ3: How would small supervision of human highlight help?}} 
We conducted experiments to investigate how our model behaves when different percentages of human-annotated data are included in the training set. 
\cref{fig:cose-portion} showcases the outcomes obtained for all training criteria when varying percentages of human annotation were used: 0.1\%, 1\%, 10\%, 20\%, 50\%, and 100\%. 
The results indicate that until 10\% of the data is annotated by humans, the plausibility remains consistent. 
On the other hand, \textsc{Refer} achieves comparable plausibility to 100\% human supervision with just 50\% of human annotation. 
This means \textsc{Refer} enables effective plausibility optimizations using minimal gold rationale supervision.
In contrast, task performance is reduced by increasing the human rationale supervision since the model should learn from human highlights instead of repetitive patterns. 
Faithfulness does not exhibit a clear relationship with the availability of gold rationales, as it relies on the model's intrinsic features rather than human-provided rationales.

\begin{table}[t]
\centering
\caption{Comparison of the performance of \textsc{Refer} without explanation regularization on ID and OOD dataset.}
\label{tab:baseline}
\resizebox{\columnwidth}{!}{%
\begin{tabular}{lcccccccc}
\hline
\multicolumn{1}{c}{\multirow{2}{*}{\textbf{Metrics}}} &
  \textbf{\begin{tabular}[c]{@{}c@{}}ID without ER\\ regularization\end{tabular}} &
  \textbf{} &
  \multicolumn{3}{c}{\textbf{OOD Datasets}} &
  \textbf{} &
  \multicolumn{2}{c}{\textbf{Contrast Test}} \\ \cline{2-2} \cline{4-6} \cline{8-9} 
\multicolumn{1}{c}{}       & e-SNLI &  & MNLI  & HANS  & e-MNLI &  & MNLI-Contrast & MNLI-Original \\ \midrule
Task Accuracy (↑)          & 90.47  &  & 74.65 & 67.09 & 76.00  &  & 82.66         & 88.72         \\
Task Macro F1 (↑)          & 90.48  &  & 74.80 & 28.57 & 75.93  &  & 60.25         & 88.74         \\
Sufficiency AOPC (↓)       & 0.205  &  & 0.206 & 0.305 & 0.249  &  & 0.226         & 0.201         \\
Comprehensiveness AOPC (↑) & 0.243  &  & 0.212 & 0.272 & 0.224  &  & 0.210         & 0.249         \\
Plausibility TF1 (↑)       & 0.254  &  & N/A   & N/A   & 0.197  &  & N/A           & N/A           \\
Plausibility AUPRC (↑)     & 0.211  &  & N/A   & N/A   & 0.167  &  & N/A           & N/A           \\ \hline
\end{tabular}%
}
\end{table}
\begin{table}[t!]
\centering
\caption{Comparison of the performance of \textsc{Refer} with explanation regularization on ID and OOD dataset.}
\label{tab:ood}
\resizebox{\columnwidth}{!}{%
\begin{tabular}{lcccccccc}
\hline
\multicolumn{1}{c}{\multirow{2}{*}{\textbf{Metrics}}} &
  \textbf{\begin{tabular}[c]{@{}c@{}}ID with ER\\ regularization\end{tabular}} &
  \textbf{} &
  \multicolumn{3}{c}{\textbf{OOD Datasets}} &
  \textbf{} &
  \multicolumn{2}{c}{\textbf{Contrast Test}} \\ \cline{2-2} \cline{4-6} \cline{8-9} 
\multicolumn{1}{c}{}       & e-SNLI &  & MNLI  & HANS  & e-MNLI &  & MNLI-Contrast & MNLI-Original \\ \midrule
Task Accuracy (↑)                                     & 90.33               &           & 74.10        & 66.06       & 78.00  &  & 82.11         & 88.37\\
Task Macro F1 (↑)                                     & 90.36               &           & 74.13        & 27.75       & 78.11  &  & 59.92         & 88.44\\
Sufficiency AOPC (↓)                                  & 0.059               &           & 0.109        & 0.071       & 0.100  &  & 0.091         & 0.050\\
Comprehensiveness AOPC (↑)                            & 0.329               &           & 0.310        & 0.320       & 0.315  &  & 0.321         & 0.329\\
Plausibility TF1 (↑)                                  & 0.792               &           & N/A          & N/A         & 0.616  &  & N/A           & N/A  \\
Plausibility AUPRC (↑)                                & 0.869               &           & N/A          & N/A         & 0.445  &  & N/A           & N/A  \\ \bottomrule
\end{tabular}
}
\end{table}
%
%
%
%
%
%
\paragraph{\textit{RQ4: Does learned rationale extractor generalize over OOD data?}}
%
%
%
\cref{tab:baseline} and \cref{tab:ood} show the \textsc{Refer} results on ID and OOD datasets. In both Tables \textsc{Refer} is trained on ID dataset and evaluated over ID and OOD sets.
We consider the results from \cref{tab:baseline} as the baseline and analyze the effect of ER regularization in \cref{tab:ood}.
When we train the model with explanation regularization, faithfulness and sufficiency are enhanced. 
On MNLI, sufficiency improves from 0.206 to 0.109, while on HANS, it goes from 0.249 to 0.071.
Regarding Comprehensiveness, training the model along with ER regularization improves the baseline from 0.212 to 0.310 on MNLI and from 0.272 to 0.320  on HANS.
Besides, results on e-MNLI in \cref{tab:ood} show that the plausibility of OOD is significant and comparable to the ID data. 
%
%
Similarly, the comprehensiveness and sufficiency improve on both MNLI-Contrast and MNLI-Original. However, the results on MNLI-Original seem to be better, especially w.r.t task macro F1, which means the model performs equally well predicting different labels.
Another interesting finding is that the model trained for a specific top-$k$\% performs well on other top-$k$\% during inference w.r.t. plausibility. \cref{fig:topk-plaus} display roughly stable behavior of the model trained for top-50\% and evaluated for other top-$k$\% w.r.t. plausibility TF1. This means the model tends to select rationales among human highlights even with a low number of $k$. \cref{tab:vary-k-highlights} illustrates the rationale selected by the model trained for top-50\% and evaluated for different $k$s.

\section{Conclusions}
In this paper, we propose \textsc{Refer}, a rationale extraction framework 
that jointly trains the task model and the rationale extractor to optimize downstream task performance, faithfulness, and plausibility. 
%
Being fully end-to-end, thanks to Adaptive Implicit Maximum Likelihood Estimation~\citep{aimle23}, enables the task model and the rationale extractor to be jointly optimized for these criteria, therefore aware of each other behavior and adopting their parameter to improve their performance and obtain a better balance.
%
We then analyze several aspects of the rationale extraction process, investigating how human rationales affect the model behavior; how the model can imitate human-generated rationales; and to what extent the learned models can generalize on OOD datasets.
Finally, by answering all these questions, we compare \textsc{Refer} performance with other methods and architectures and illustrate that our model outperforms previous models in most cases.

\paragraph{Acknowledgments}
Reza was funded by the \emph{Thesis Abroad Scholarship} from the Department of Computer Science and Engineering (DISI) at the University of Bologna.
Pasquale was partially funded by the European Union’s Horizon 2020 research and innovation program under grant agreement no. 875160, ELIAI (The Edinburgh Laboratory for Integrated Artificial Intelligence) EPSRC (grant no. EP/W002876/1), an industry grant from Cisco, and a donation from Accenture LLP, and is grateful to NVIDIA for the GPU donations. 
This work was supported by the Edinburgh International Data Facility (EIDF) and the Data-Driven Innovation Programme at the University of Edinburgh.
%


\bibliography{anthology,custom}
\bibliographystyle{acl_natbib}

\clearpage

\appendix

\section{Model Detail}
\label{sec:model-detail}
Transformers-based models, such as BERT, have been one of the most successful deep learning models for NLP. Unfortunately, one of their core limitations is the quadratic dependency (mainly in terms of memory) on the sequence length due to their full attention mechanism. To remedy this, \citet{zaheer_NEURIPS2020_c8512d14} proposed \textsc{BigBird}, a sparse attention mechanism that reduces this quadratic dependency to linear. They show that \textsc{BigBird} is a universal approximator of sequence functions and is Turing complete, thereby preserving these properties of the quadratic, full attention model. Along the way, their theoretical analysis reveals some of the benefits of having \textit{O(1)} global tokens (such as CLS) that attend to the entire sequence as part of the sparse attention mechanism. The proposed sparse attention can handle sequences of length up to eight times what was previously possible using similar hardware. Due to the capability to handle longer contexts, \textsc{BigBird} drastically improves performance on various NLP tasks such as question answering and summarization.

\section{Hyperparameters}
\label{sec:hyperparam}
In our implementation, we utilize BigBird-Base \cite{zaheer_NEURIPS2020_c8512d14} as the backbone for both \(\mathcal{F}_{\text{task}}\) and \(\mathcal{F}_{\text{ext}}\). This choice enables us to effectively handle input sequences of considerable length, accommodating up to 4096 tokens. We used AIMLE, which uses adaptive target distribution with alpha and beta initialized to 1 and 0, respectively. Throughout all experiments, we maintain a consistent learning rate of $2 \times 10^{-5}$ and employ an effective batch size of 32. Our training process spans a maximum of 10 epochs, with early stopping applied after 5 epochs of no significant improvement. 
%
%
To ensure optimal performance, we focus our hyperparameter tuning efforts on the weights associated with faithfulness and plausibility losses, specifically $\alpha_{\text{c}} = \alpha_{s} = \alpha_{\text{f}}$, and $\alpha_{\text{p}}$ as well as top-$k\%$. We applied a grid search across various configurations and evaluated their impact on comprehensiveness, sufficiency, plausibility scores, and task performance.  The entire implementation is carried out using the PyTorch-Lightning framework \cite{Paszke_NEURIPS2019_bdbca288, falcon2019pytorchlight}, which provides a streamlined and user-friendly environment for deep learning experiments.
\begin{figure}[t]
    \centering
    \includegraphics[width=\columnwidth]{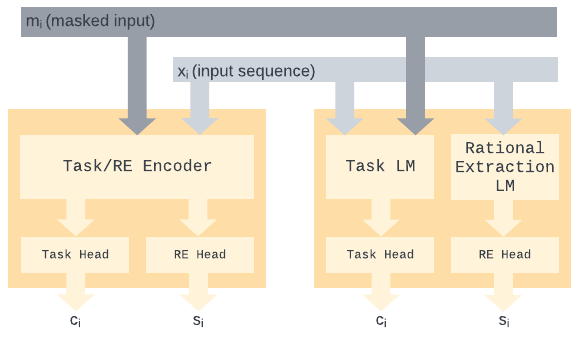}
    \label{fig:three sin x}
    \caption{Shared LM (left) and Dual LM (right) architecture. Using shared LM, the task model and rational extractor share the same encoder. While in the Dual LM model, they are completely separate}
    \label{fig:slm-dlm}
\end{figure}
\begin{table}[t]
\centering
\caption{Examples of highlights differing in comprehensiveness and sufficiency}
\label{tab:highlight-examples}
\resizebox{\columnwidth}{!}{%
\begin{tabular}{@{}ll@{}}
\toprule
\multicolumn{1}{c}{\textbf{Instance with Highlight}} &
  \multicolumn{1}{c}{\textbf{Type of Highlight}} \\ \midrule
\begin{tabular}[c]{@{}l@{}}\textbf{\textit{Premise}}: People are\hlc[yellow!30]{stretching on yoga mats.}\\ \textbf{\textit{Hypothesis}}: They\hlc[yellow!30]{stretched on bikes.}\\ \textbf{\textit{Label}}: contradiction\end{tabular} &
  \begin{tabular}[c]{@{}l@{}}\textbf{\textit{Premise}}:People are\hlc[black!50]{stretching on yoga mats.}\\ \textbf{\textit{Hypothesis}}:They\hlc[black!50]{stretched on bikes.}\\ \textcolor{teal}{(sufficient)}\end{tabular} \\
  &
  \\
\begin{tabular}[c]{@{}l@{}}\textbf{\textit{Premise}}:\hlc[yellow!30]{People on bicycles}waiting at an intersection.\\ \textbf{\textit{Hypothesis}}: There are\hlc[yellow!30]{people on bicycles.}\\ \textbf{\textit{Label}}: entailment\end{tabular} &
  \begin{tabular}[c]{@{}l@{}}\textbf{\textit{Premise}}:\hlc[black!50]{People on bicycles}waiting at an intersection.\\ \textbf{\textit{Hypothesis}}:There are\hlc[black!50]{people on bicycles.}\\ \textcolor{teal}{(comprehensive)}\end{tabular} \\
  &
  \\
\begin{tabular}[c]{@{}l@{}}\textbf{\textit{Premise}}: People on bicycles waiting at an intersection.\\ \textbf{\textit{Hypothesis}}: Some\hlc[yellow!30]{people}on bikes are\hlc[yellow!30]{stopped at a junction.}\\ \textbf{\textit{Label}}: neutral\end{tabular} &
  \begin{tabular}[c]{@{}l@{}}\textbf{\textit{Premise}}:\hlc[black!50]{People on bicycles waiting at an intersection.}\\ \textbf{\textit{Hypothesis}}:\hlc[black!50]{Some}people\hlc[black!50]{on bikes are}stopped at a junction.\\ \textcolor{red}{(¬ sufficient)}\end{tabular} \\ \bottomrule
\end{tabular}%
}
\end{table}
\section{OOD Generalization}

\begin{figure}[t]
    \centering
    \includegraphics[width=\columnwidth]{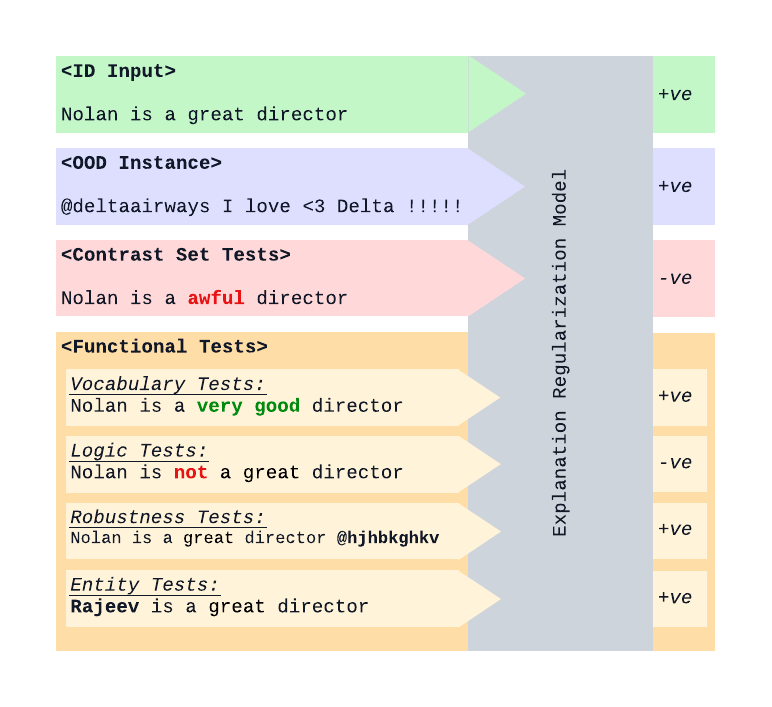}
    \caption{ER-TEST Framework - Apart from existing ID evaluations of ER criteria, ER-TEST evaluates ER’s impact on OOD generalization along three dimensions: A. Unseen datasets, B. Contrast set tests, and C. Functional tests.} 
    \label{fig:er-test}
\end{figure}

\begin{table}[t]
\centering
\caption{e-MNLI instances for different labels. Following e-SNLI for neutral labels only tokens in hypothesis are highlighted.}
\label{tab:emnli}
\resizebox{\columnwidth}{!}{%
\begin{tabular}{@{}lc@{}}
\toprule
\multicolumn{1}{c}{\textbf{Instances with Highlights}}           & \multicolumn{1}{c}{\textbf{\textit{Label}}} \\ \midrule
\begin{tabular}[c]{@{}l@{}}\textbf{Premise}: They drive it around the country in a dilapidated ice-cream truck trying to\hlc[yellow!30]{keep it cool.}\\ \textbf{\textit{Hypothesis}}: They used an ice cream truck to try and\hlc[yellow!30]{keep it from getting warm.}\end{tabular} & entailment                                              \\
                                                                 &
                                                        \\
\begin{tabular}[c]{@{}l@{}}\textbf{\textit{Premise}}: Then he turned to Tommy.\\ \textbf{\textit{Hypothesis}}: He\hlc[yellow!30]{talked}to Tommy.\end{tabular} & neutral                           \\
                                                                 &  
                                                        \\
\begin{tabular}[c]{@{}l@{}}\textbf{\textit{Premise}}: but i've\hlc[yellow!30]{lived up here all my life}and i'm fifty eight years old so i i could\\ \textbf{\textit{Hypothesis}}: I have\hlc[yellow!30]{moved somewhere else}in my life.\end{tabular} & contradiction                      \\ \bottomrule
\end{tabular}
}
\end{table}

Out-of-distribution (OOD) generalization refers to the ability of a model to accurately handle data samples that deviate from the distribution of its training data.
OOD generalization is a critical challenge in NLP tasks and plays a pivotal role in ensuring the reliability and effectiveness of NLP models in real-world applications.
%
Effective OOD generalization in NLP requires models to capture and understand the underlying linguistic properties and generalizable patterns rather than relying on memorization or overfitting specific training instances. 
However, despite the growing interest in OOD generalization, existing evaluations in the field of explanation robustness have been limited in scope and coverage. 
Existing works primarily evaluate explanation regularization models via in-distribution (ID) generalization \cite{zaidan-etal-2007-using,lin-etal-2020-triggerner,huang-etal-2021-exploring}, though a small number of works have done auxiliary evaluations of OOD generalization \cite{ross2017right,kennedy-etal-2020-contextualizing,rieger2020interpretations}. 
Consequently, there is a lack of comprehensive understanding regarding the impact of explanation robustness on OOD generalization. 
To address this gap,  \citet{joshi-etal-2022-er} introduce ER-TEST, a unified benchmark specifically designed to assess the OOD generalization capabilities of explanation regularization models across three dimensions. 
These dimensions include evaluating models on (i) unseen datasets, (ii) conducting contrast set tests to measure their ability to handle diverse and challenging inputs, and (iii) functional tests which include four scopes: vocabulary tests, logic tests, robustness tests, and entity tests -- the functional test is not included in our work. We leave this field for future work -- to assess their reasoning and inference capabilities. Examples of each dimension are shown in \cref{fig:er-test}.
Ideally, we would like the explanation regularization model to perform well on all three aspects during the evaluation of OOD data. 
However, since the datasets for OOD evaluation do not contain human-annotated rationales there is no possibility of assessing the plausibility criteria. 
%
%
By addressing the OOD generalization challenge, NLP models can achieve greater robustness, adaptability, and practical utility in real-world scenarios, thus advancing the field of natural language processing and
can better handle challenging scenarios.

\begin{table}[t]
\centering
\caption{The heuristics targeted by the HANS dataset, along with examples of incorrect entailment predictions that these heuristics would lead to.}
\label{tab:huristics}
\resizebox{\columnwidth}{!}{%
\begin{tabular}{@{}lll@{}}
\toprule
\multicolumn{1}{c}{\textbf{Heuristic}} & \multicolumn{1}{c}{\textbf{Definition}} & \multicolumn{1}{c}{\textbf{Example}} \\ \midrule
Lexical overlap &
  \begin{tabular}[c]{@{}l@{}}The premise entails all hypotheses \\ constructed from its own words.\end{tabular} &
  \begin{tabular}[c]{@{}l@{}}The \textbf{judges} \textbf{admired} the \textbf{doctors}.\\ $\xrightarrow{\text{Wrong}}$ The \textbf{doctors} \textbf{admired} the \textbf{judges} .\end{tabular} \\
  &  &  \\
Subsequence &
  \begin{tabular}[c]{@{}l@{}}The premise entails all of its \\ contiguous subsequences.\end{tabular} &
  \begin{tabular}[c]{@{}l@{}}\textbf{The lawyers believed the bankers} resigned.\\ $\xrightarrow{\text{Wrong}}$ The lawyers believed the bankers.\end{tabular} \\
  &  &  \\
Constituent &
  \begin{tabular}[c]{@{}l@{}}The premise entails all complete \\ subtrees in its parse tree.\end{tabular} &
  \begin{tabular}[c]{@{}l@{}}Probably \textbf{the tourists waited}.\\ $\xrightarrow{\text{Wrong}}$ The tourists waited.\end{tabular} \\ \bottomrule
\end{tabular}%
}
\end{table}

\begin{table}[t!]
\centering
\caption{Comparison of rationales extracted by \textsc{Refer} trained on $k$=50\%. We forced the model for other $k$ to see how it selects rationales.}
\label{tab:vary-k-highlights}
\resizebox{\columnwidth}{!}{%
\begin{tabular}{@{}cl@{}}
\toprule
\textbf{Dataset} & \multicolumn{1}{c}{\textbf{Test Instance}}                                                                                                                                 \\ \midrule
Gold     & \begin{tabular}[c]{@{}l@{}}\textbf{\textit{Premise}}: a woman wearing a\hlc[yellow!30]{pink tank top holding a mug of liquid}\\ \textbf{\textit{Hypothesis}}: A woman in a\hlc[yellow!30]{blue tank top holding a car.}\\ \textbf{\textit{Label}}: contradiction\end{tabular} \\
               &                                                                                                                                                                            \\
k=20\%          & \begin{tabular}[c]{@{}l@{}} \textbf{\textit{Premise}}: a woman wearing a\hlc[orange!60]{pink}tank top holding a\hlc[orange!60]{mug}of liquid\\ \textbf{\textit{Hypothesis}}: A woman in a\hlc[orange!60]{blue}tank top holding a\hlc[orange!60]{car.}\end{tabular}                                                                                                            \\
               &                                                                                                                                                                            \\
k=30\%          & \begin{tabular}[c]{@{}l@{}} \textbf{\textit{Premise}}: a woman wearing a\hlc[orange!60]{pink}tank top holding a\hlc[orange!60]{mug of liquid}\\ \textbf{\textit{Hypothesis}}: A woman in a\hlc[orange!60]{blue}tank top holding a\hlc[orange!60]{car.}\end{tabular}                                                                                                            \\
               &                                                                                                                                                                            \\
k=40\%          & \begin{tabular}[c]{@{}l@{}}\textbf{\textit{Premise}}: a woman wearing a\hlc[orange!60]{pink tank top}holding a\hlc[orange!60]{mug of liquid}\\ \textbf{\textit{Hypothesis}}:A woman in a\hlc[orange!60]{blue tank}top holding a\hlc[orange!60]{car.}\end{tabular}                                                                                                            \\
               &                                                                                                                                                                            \\
k=50\%          & \begin{tabular}[c]{@{}l@{}}\textbf{\textit{Premise}}: a woman wearing a\hlc[orange!60]{pink tank top}holding a\hlc[orange!60]{mug of liquid}\\ \textbf{\textit{Hypothesis}}: A woman in a\hlc[orange!60]{blue tank top holding}a\hlc[orange!60]{car.}\end{tabular}                                                                                                            \\
               &                                                                                                                                                                            \\
k=60\%          & \begin{tabular}[c]{@{}l@{}}\textbf{\textit{Premise}}: a woman wearing a\hlc[orange!60]{pink tank top holding}a\hlc[orange!60]{mug of liquid}\\ \textbf{\textit{Hypothesis}}: A woman in a\hlc[orange!60]{blue tank top holding a car.}\end{tabular}                                                                                                            \\ \bottomrule
\end{tabular}
}
\end{table}
\begin{table}[t!]
\centering
\caption{MNLI Contrast Test Set. In the MNLI-Original the original label is unchanged while in the MNLI-Contrast the label is also changed based on changes in premise or hypothesis.}
\label{tab:mnli-contrast}
\resizebox{\columnwidth}{!}{%
\begin{tabular}{@{}cl@{}}
\toprule
\textbf{Model} & \multicolumn{1}{c}{\textbf{Contrast Set Instance}}                       \\ \midrule
MNLI-Contrast  & 
\begin{tabular}[c]{@{}l@{}}
\textit{\textbf{Premise}}: yeah well that's not really immigration.\\
$\xrightarrow[]{\text{past simple}}$ Yeah well that wasn't immigration.\\ 
\textbf{\textit{Hypothesis}}: That is not immigration.\\
$\xrightarrow[]{\text{future simple}}$ That won't be immigration.\\
\textbf{\textit{Label}}: entail$\xrightarrow{}$ neutral
\end{tabular} 
\\ \midrule
MNLI-Original  & 
\begin{tabular}[c]{@{}l@{}}
\textbf{\textit{Premise}}: Clearly, GAO needs assistance to meet its \\looming human capital challenges.\\$\xrightarrow[]{\text{it cleft: ARG1}}$ Clearly it is GAO who needs assistance \\to meet its human capital challenges looming.\\
\textbf{\textit{Hypothesis}}: GAO will soon be suffering from a shortage \\of qualified personnel.\\$\xrightarrow[]{\text{it cleft: ARG1}}$ It is GAO who soon will be suffering from a \\shortage of personnel qualified for.\\
\textbf{\textit{Label}}: neutral$\xrightarrow{}$ neutral
\end{tabular} \\ \bottomrule
\end{tabular}%
}
\end{table}



\begin{table*}[t!]
\centering
\caption{Benchmark on CoS-E dataset. Results of the baselines are obtained from the work done by \citet{chan-etal-2022-unirex}.}
\label{tab:cose-results}
\resizebox{\textwidth}{!}{%
\begin{tabular}{@{}ccccccccccccccc@{}}
\toprule
\multicolumn{2}{c}{Configuration} &
   &
  \multicolumn{3}{c}{Faithfulness} &
   &
  \multicolumn{3}{c}{Plausibility} &
   &
  \multicolumn{2}{c}{Task} &
   &
  \multicolumn{1}{c}{Composite} \\ \cmidrule(r){1-2} \cmidrule(lr){4-6} \cmidrule(lr){8-10} \cmidrule(lr){12-13} \cmidrule(l){15-15} 
Model &
  End-to-End &
   &
  Comp (↑) &
  Suff (↓) &
  FNRG &
   &
  TF1 (↑) &
  AUPRC (↑) &
  PNRG &
   &
  Accuracy (↑) &
  TNRG &
   &
  \multicolumn{1}{c}{CNRG} \\ \midrule
AA(IG)          & FALSE &  & 0.2160 & 0.3780 & 0.3306 &  & 0.4834 & 0.4007 & 0.2935 &  & 63.56 & 0.9772 &  & 0.5337 \\
SGT             & FALSE &  & 0.1970 & 0.3240 & 0.3699 &  & 0.5100 & 0.4368 & 0.3702 &  & 64.35 & 0.9950 &  & 0.5783 \\
FRESH           & FALSE &  & 0.0370 & 0.0000 & 0.5463 &  & 0.3937 & 0.3235 & 0.0849 &  & 24.81 & 0.1007 &  & 0.2439 \\
A2R             & FALSE &  & 0.0140 & 0.0000 & 0.5167 &  & 0.3312 & 0.4161 & 0.1041 &  & 21.77 & 0.0319 &  & 0.2176 \\
SGT+P           & FALSE &  & 0.2010 & 0.3280 & 0.3703 &  & 0.4795 & 0.413  & 0.3020 &  & \textbf{64.57} & \textbf{1.0000} &  & 0.5574 \\
FRESH+P         & FALSE &  & 0.0130 & 0.0130 & 0.5001 &  & 0.6976 & 0.7607 & 0.9890 &  & 20.36 & 0.0000 &  & 0.4964 \\
A2R+P           & FALSE &  & 0.0010 & 0.0000 & 0.5000 &  & 0.6763 & 0.7359 & 0.9322 &  & 20.91 & 0.0124 &  & 0.4816 \\ \midrule
UNIREX (DLM+P)  & FALSE &  & 0.1800 & 0.3900 & 0.2702 &  & 0.6976 & 0.7607 & 0.9890 &  & 64.13 & 0.9900 &  & 0.7497 \\
UNIREX (DLM+FP) & FALSE &  & 0.2930 & 0.3210 & 0.4968 &  & 0.6952 & 0.7638 & 0.9892 &  & 62.5  & 0.9532 &  & \textbf{0.8131} \\
UNIREX (SLM+FP) & FALSE &  & 0.3900 & 0.4240 & 0.5000 &  & 0.6925 & 0.7512 & 0.9714 &  & 62.09 & 0.9439 &  & 0.8051 \\ \midrule
REFER+P         & TRUE  &  & 0.1831 & 0.2098 & 0.4867 &  & \textbf{0.6994} & \textbf{0.7683} & \textbf{1.0000} &  & 61.35 & 0.9272 &  & 0.8046 \\
REFER+F         & TRUE  &  & \textbf{0.2798} & \textbf{0.0000} & \textbf{0.8584} &  & 0.3835 & 0.6691 & 0.4595 &  & 63.21 & 0.9692 &  & 0.7624 \\
REFER+FP        & TRUE  &  & 0.1206 & 0.1489 & 0.4781 &  & 0.6881 & 0.7393 & 0.9521 &  & 64.23 & 0.9923 &  & 0.8075 \\ \bottomrule
\end{tabular}%
}
\end{table*}

\begin{table*}[t]
\centering
\caption{Benchmark on e-SNLI dataset. Results of the baselines are obtained from the work done by \citet{chan-etal-2022-unirex}.}
\label{tab:esnli-results}
\resizebox{\textwidth}{!}{%
\begin{tabular}{@{}ccccccccccccccc@{}}
\toprule
\multicolumn{2}{c}{Configuration} &
   &
  \multicolumn{3}{c}{Faithfulness} &
   &
  \multicolumn{3}{c}{Plausibility} &
   &
  \multicolumn{2}{c}{Task} &
   &
  \multicolumn{1}{c}{Composite} \\ \cmidrule(r){1-2} \cmidrule(lr){4-6} \cmidrule(lr){8-10} \cmidrule(lr){12-13} \cmidrule(l){15-15} 
Model &
  End-to-End &
   &
  Comp (↑) &
  Suff (↓) &
  FNRG &
   &
  TF1 (↑) &
  AUPRC (↑) &
  PNRG &
   &
  Macro F1 (↑) &
  TNRG &
   &
  \multicolumn{1}{c}{CNRG} \\ \midrule
AA(IG)          & FALSE &  & 0.3080 & 0.4140 & 0.4250 &  & 0.3787 & 0.4783 & 0.1728 &  & 90.78 & 0.9909 &  & 0.5296 \\
SGT             & FALSE &  & 0.2880 & 0.3610 & 0.4557 &  & 0.4170 & 0.4246 & 0.1551 &  & 90.23 & 0.9766 &  & 0.5291 \\
FRESH           & FALSE &  & 0.1200 & 0.0000 & 0.6117 &  & 0.5371 & 0.3877 & 0.2337 &  & 72.92 & 0.5259 &  & 0.4571 \\
A2R             & FALSE &  & 0.0530 & 0.0000 & 0.5000 &  & 0.2954 & 0.4848 & 0.0989 &  & 52.72 & 0.0000 &  & 0.1996 \\
SGT+P           & FALSE &  & 0.2860 & 0.3390 & 0.4789 &  & 0.4259 & 0.4303 & 0.1696 &  & 90.36 & 0.9800 &  & 0.5428 \\
FRESH+P         & FALSE &  & 0.1430 & 0.0000 & 0.6500 &  & 0.7763 & 0.8785 & 0.9649 &  & 73.44 & 0.5394 &  & 0.7181 \\
A2R+P           & FALSE &  & 0.1820 & 0.0000 & 0.7150 &  & 0.7731 & 0.873  & 0.9562 &  & 77.31 & 0.6402 &  & 0.7705 \\ \midrule
UNIREX (DLM+P)  & FALSE &  & 0.3110 & 0.3710 & 0.4819 &  & 0.7763 & 0.8785 & 0.9649 &  & 90.8  & 0.9914 &  & 0.8127 \\
UNIREX (DLM+FP) & FALSE &  & 0.3350 & 0.3460 & 0.5521 &  & 0.7753 & 0.8699 & 0.9552 &  & 90.51 & 0.9839 &  & 0.8304 \\
UNIREX (SLM+FP) & FALSE &  & 0.3530 & 0.3560 & 0.5700 &  & 0.7722 & 0.8758 & 0.9582 &  & 90.59 & 0.9859 &  & 0.8381 \\ \midrule
REFER+P         & TRUE  &  & 0.3127 & 0.1768 & 0.7193 &  & 0.7909 & 0.8411 & 0.9409 &  & 87.81 & 0.9136 &  & 0.8579 \\
REFER+F         & TRUE  &  & \textbf{0.3054} & \textbf{0.0000} & \textbf{0.9207} &  & 0.4443 & 0.5958 & 0.3559 &  & 90.69 & 0.9885 &  & 0.7551 \\
REFER+FP        & TRUE  &  & 0.3091 & 0.0399 & 0.8786 &  & \textbf{0.8126} & \textbf{0.8713} & \textbf{0.9927} &  & \textbf{91.13} & \textbf{1.0000} &  & \textbf{0.9571} \\ \bottomrule
\end{tabular}%
}
\end{table*}

\end{document}